\documentclass[letterpaper, 10 pt, conference]{ieeeconf}  

\IEEEoverridecommandlockouts                              

\usepackage{epsfig}
\usepackage{makecell}
\usepackage{subcaption}
\usepackage{xcolor}
\usepackage{ragged2e}
\usepackage{float} 

\usepackage{graphicx}
\usepackage{amsmath}
\usepackage{adjustbox}
\usepackage{amssymb}
\usepackage{booktabs}
\usepackage{tabularx}
\usepackage{array}
\usepackage{makecell}
\usepackage{bm}

\title{\LARGE \bf
GPS-Enhanced Tourist Mobility Modeling with\\
Seasonal Spatial Priors and LLM-Based Activity Chain Generation
}

\author{Yifan Liu$^{1,*}$, Yanling Sang$^{1,*}$, Xishun Liao$^{3}$, Morgan Sun$^{2}$, Bo Yang$^{1}$, Zhiyuan Zhang$^{1}$,\\
Chris Stanford$^{2}$, Haoxuan Ma$^{1,\dagger}$, and Jiaqi Ma$^{1}$
\thanks{$^{*}$Equal contribution.}
\thanks{$^{1}$UCLA Mobility Lab, Department of Civil and Environmental Engineering, University of California, Los Angeles, Los Angeles, USA.}
\thanks{$^{2}$Novateur Research Solutions, Ashburn, VA, USA.}
\thanks{$^{3}$University of Central Florida, Orlando, FL, USA.}
\thanks{$^{\dagger}$Corresponding author: haoxuanma@ucla.edu}
}

\begin{document}

\maketitle
\thispagestyle{empty}
\pagestyle{empty}

\begin{abstract}
Tourist mobility poses a distinct challenge for urban transportation planning. Unlike resident commuting, tourist travel is largely non-routine, attraction driven, and highly sensitive to trip purpose, travel season, and trip member composition. Existing approaches either measure aggregate tourist spatial patterns without generating individual schedules, or synthesize mobility without tourist specific structure such as trip duration conditioning, month varying attraction demand, and household co-travel rules. To address these challenges, we propose a four stage simulation framework combining month conditioned spatial priors derived from GPS and survey data, trip extent prediction from tourist demographics, distance feasible ward sequence assignment, and LLM-based activity chain generation under household and spatial constraints. GPS data are used only in privacy preserving aggregated form as month conditioned spatial priors, with no individual traces retained or exposed. Experiments on tourism in Tokyo demonstrate that the GPS based tourist cohort extraction recovers spatial visitation signatures consistent with survey references, and our framework produces demographically aligned synthetic schedules whose ward-level visitation shares align closely with both survey distributions and staypoint derived monthly visitation patterns. The results demonstrate the framework's effectiveness as a geographically grounded, demographically aware approach to tourist mobility modeling.
\end{abstract}

\section{INTRODUCTION}

Modeling traveler behavior with high fidelity is a fundamental undertaking in transportation planning, influencing transit operations, infrastructure development, and crowd management strategies~\cite{bowman2001activity, pitakaso2025fuzzy, signorile2018mobility, 11131172}. For cities that attract large volumes of visitors, this challenge takes on an additional dimension. The structural characteristics of traveler mobility differ markedly from those of residents, being shaped by attraction popularity, seasonal patterns, and trip purpose rather than habitual commuting behavior~\cite{lew2006modeling, liyanage2025demand}. Yet most activity based and mobility simulation frameworks are calibrated for residents, leaving a persistent gap in planning grade models for short stay visitor populations~\cite{ma2025learning}.

Tourist mobility has several properties that render standard modeling assumptions invalid. Tourists do not follow the home–work or other routine patterns typical of residents~\cite{jiang2024investigating}. Tourists' destination choices are attraction driven and shift with travel month and POI popularity, with demand volumes and spatial patterns varying substantially across the year~\cite{jin2018using}. Trip horizons are short, with itinerary decisions often made without any prior knowledge of the destination, leaving no historical behavioral trace for individual level models to learn from~\cite{senefonte2022predictour}. Moreover, travel is frequently organized by households with purpose dependent role divergence: a business conference attendee and their companion follow different daytime itineraries within the same group~\cite{cai2001profiling}. These properties make tourist mobility uniquely sparse, variable, and context driven, necessitating tourist specific modeling rather than adapting resident oriented tools.

Existing approaches can be grouped into three families, each addressing part of this problem but leaving a critical gap for end-to-end tourist specific demand synthesis. Passive sensing and flow network studies have substantially advanced our ability to measure where tourists go and when~\cite{shoval2007tracking}, but remain focused on aggregate measurement and stop short of generating individual schedules. Sequence prediction and trajectory generation models achieve strong next location accuracy on dense check-in histories~\cite{crivellari2020lstm}, but this setting is unavailable for short-stay tourists arriving in an unfamiliar city with no prior trace. LLM-based mobility generators show semantic plausibility for resident activity chains~\cite{wang2024llmob}, but were not designed with tourism specific structure: trip-duration conditioning, seasonal attraction demand, and household co-travel rules are absent.

Building on these lines of research, we introduce a four stage tourist mobility modeling framework. Stage 0 extracts a tourist cohort from anonymous GPS staypoints and derives month conditioned ward visitation priors. Stage 1 predicts per traveler trip extent from tourist demographics via gradient boosting~\cite{chen2016xgboost}, conditioned on a synthetic base population drawn from survey marginals. Stage 2 constructs distance feasible daily ward sequences calibrated to seasonal attraction demand. Stage 3 generates quarter-hour activity chains via an LLM under explicit spatial, temporal, and household co-travel constraints, with purpose dependent companion variation. The framework fuses survey demographic attributes, GPS monthly aggregates, and inter-ward distance matrices to produce planning grade synthetic tourist schedules without exposing individual GPS traces. Our study makes several key contributions to the field of tourist mobility modeling compared with existing approaches:

\begin{figure*}[t]
    \centering
    \includegraphics[width=\textwidth]{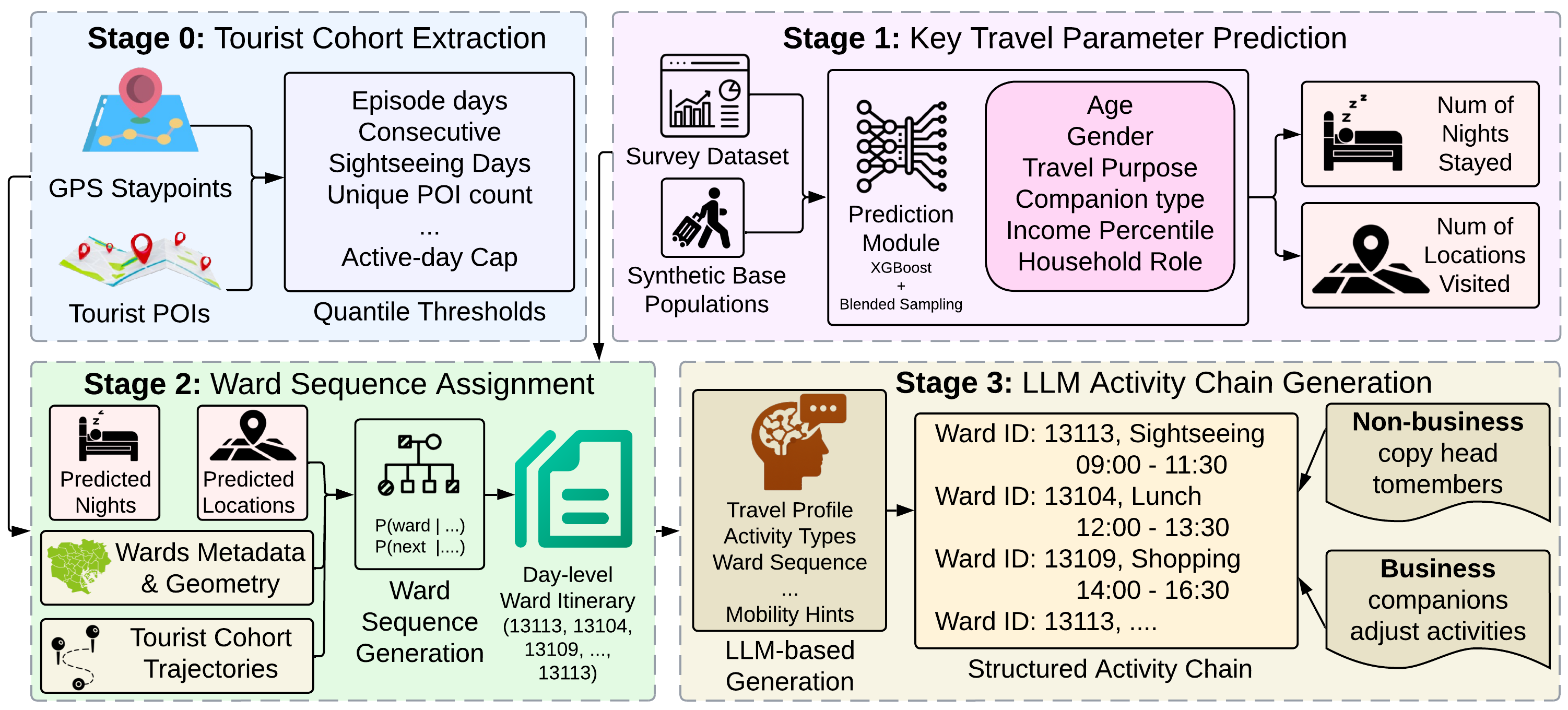}
    \caption{Overview of the proposed four stage tourist mobility modeling framework for tourist itinerary generation.}
    \label{fig:pipeline}
\end{figure*}

\begin{itemize}
    \item We propose a four stage tourist mobility modeling framework that produces demographic aware, spatially grounded synthetic tourist itineraries without requiring individual level GPS traces.

    \item We introduce a tourist cohort extraction algorithm using anonymous GPS staypoints, and derive month conditioned ward visitation and transition priors from the cohort, enabling synthetic schedules to capture seasonal attraction demand in a privacy preserving manner.

    \item We develop an LLM-based activity chain module under explicit household co-travel and spatial constraints, producing demographically consistent, temporally coherent tourist schedules at planning fidelity.
\end{itemize}

\section{LITERATURE REVIEW}

\subsection{Tourist mobility measurement and tracking}
\label{sec:lit_tracking}
Tourism research has long sought high resolution space time observation of visitor behavior. Foundational work on tourist movement patterns showed that spatial itineraries within a destination are shaped by attraction rank, distance, and trip length characteristics \cite{rossello2005modeling, gomez2025advanced}. Building on this, digital tracking studies established the value of GPS and mobile data for capturing tourist movement paths and dwell timing \cite{zheng2017understanding}, with subsequent work revealing temporal heterogeneity in tourist flow networks across different trip lengths \cite{raun2016measuring}. At larger scale, mobile positioning has been shown to support official tourism statistics including inbound and outbound flows \cite{saluveer2020methodological}. Deep learning has further extended these capabilities: LSTM based models trained on large-scale anonymized CDR data have shown that the non-repetitive nature of tourist traces favors collective learning over individual histories \cite{crivellari2020lstm}. Network based methods have further characterized city-scale tourism flow structure using social media and cellular data \cite{wang2021nanjing}. A recent synthesis of a decade of big data research on tourist mobility highlights that while aggregate sensing has matured substantially, individual level itinerary modeling remains an open challenge \cite{chen2024tracking}. Together, these studies establish the empirical value of passively sensed data, but target aggregate measurement or single step prediction rather than full individual level schedule generation.

\subsection{Data driven mobility and next-location modeling}
\label{sec:lit_dl}
The machine learning community has advanced next-location and next-POI prediction through increasingly expressive sequence models. DeepMove introduced attentional recurrent networks with periodicity modeling for next-location prediction \cite{feng2018deepmove}. Flashback improved sparse trace modeling through context aware retrieval of recurrent hidden states \cite{yang2020flashback}, and GETNext further leveraged a collective trajectory flow graph with a Transformer encoder for next-POI recommendation \cite{yang2022getnext}. On the generative side, TrajGAIL applied generative adversarial imitation learning to produce urban vehicle trajectories with distributional fidelity \cite{choi2021trajgail}, while Traveller demonstrated travel pattern awared generation via autoregressive diffusion models \cite{luo2026traveller}. Despite their predictive power, these methods assume a persistent behavioral history per user and do not incorporate trip duration conditioning, month varying attraction demand, or household co-travel structure, all essential for the tourist setting.

\subsection{LLM-based mobility generation}
\label{sec:lit_llm}
Large language models have rapidly expanded from next-token prediction to behavior simulation and schedule generation. Recent research has framed LLM agents for personal mobility generation with alignment and retrieval components \cite{munir2025pedestrian, liu2024human, gao2026foundation}. CoPB introduced a psychologically inspired reasoning workflow that combines LLMs with mechanistic models to improve intention modeling while reducing token cost \cite{shao2024chain}. TrajLLM proposed a modular agent based trajectory simulation architecture \cite{ju2025trajllm}, and AgentMove demonstrated zero-shot next-location prediction through agentic task decomposition \cite{feng2025agentmove}. These studies confirm the semantic richness that LLMs bring to mobility modeling, but remain focused on generic resident mobility or single-step prediction without tourism specific trip length conditioning, month conditioned spatial priors, or household co-travel constraints.

Our framework sits at the intersection of the above three areas. Unlike tracking studies, we target individual level schedule generation rather than aggregate measurement. Unlike sequence models, we do not require behavioral histories, instead conditioning on demographics, month varying spatial priors, and household context. Unlike LLM mobility generators, we embed tourism specific trip-scope prediction and ward-level routing, grounding generation in GPS derived spatial evidence.

\section{Methodology}

\subsection{Problem Formulation}

To formulate our problem, consider a tourist agent $i$ with their socio-demographic profile information
$\mathbf{D}_i=(\text{age}_i,\text{gender}_i,\text{purpose}_i,\text{companion}_i,\ldots,\text{household\ role}_i)$,
the objective is to generate a multi-day activity chain for that agent. Each activity episode is represented by activity type, start time, end time, and visited ward:
\begin{equation}
\mathbf{C}_i=\left\{[a_{ik},t^s_{ik},t^e_{ik},w_{ik}]\right\}_{k=1}^{K_i}.
\label{eq:chain_def}
\end{equation}
Here, $a_{ik}$ is activity type, $t^s_{ik}$ and $t^e_{ik}$ are quarter-hour start/end indices, and $w_{ik}$ is ward ID.

At the framework level, we learn a mapping from demographics to structured schedules through intermediate travel scope and spatial decisions:
\begin{equation}
\mathcal{F}: \mathbf{D}_i \rightarrow (\hat{p}^{\text{nights}}_i,\hat{p}^{\text{locs}}_i,\hat{\mathbf{W}}_i,\hat{\mathbf{C}}_i),
\label{eq:problem_map}
\end{equation}
where $\hat{p}^{\text{nights}}_i$ is the predicted number of nights stayed in Tokyo, $\hat{p}^{\text{locs}}_i$ is the predicted number of visited destinations, $\hat{\mathbf{W}}_i$ is the day-level ward sequence, and $\hat{\mathbf{C}}_i$ is the final activity chain.

\subsection{Framework Overview}

Figure~\ref{fig:pipeline} illustrates the four stage framework.
The framework takes two prerequisite inputs: a \emph{synthetic base population} drawn from Tokyo tourist survey marginals providing per agent demographic and trip attributes, and anonymous large-scale GPS staypoint data.
Stage 0 extracts a high confidence tourist cohort from those staypoints and derives month conditioned ward visitation priors.
Stage 1 predicts per traveler trip scope including nights and number of locations via multi-class gradient boosting applied to the synthetic population.
Stage 2 constructs quota calibrated and distance feasible daily ward sequences using survey marginals and GPS derived month conditioned transition priors.
Stage 3 generates quarter hour activity chains using an LLM conditioned on ward sequences and household context. Throughout the framework, GPS trajectories are utilized solely at an aggregate level, either to provide prior information or serve as an independent validation reference, ensuring complete privacy preservation. At no stage are individual level traces used directly for itinerary generation.

\subsection{Stage 0: GPS Tourist Cohort Extraction}

Stage 0 derives a high confidence tourist cohort and month conditioned spatial priors from raw GPS staypoints via rule based POI proximity matching. The POI catalog contains 60 entries and was curated to cover both leisure attractions and business relevant anchors across Tokyo (temples/shrines, museums, shopping areas, station hubs, business districts, conference venues, airport nodes). As shown in Table~\ref{tab:poi_examples}, each POI record stores centroid coordinates geocoded from OpenStreetMap, category, match radius $r_p$ set proportional to the site's physical footprint and category level GPS positioning uncertainty, and minimum dwell threshold $\delta_p$ reflecting meaningful engagement over pass through.

\begin{table}[h]
    \centering
    \caption{Example POIs used in Stage~0 tourist extraction.}
    \label{tab:poi_examples}
    \scriptsize
    \begin{tabular}{lccc}
    \toprule
    POI & Category & $r_p$ (m) & $\delta_p$ (s) \\
    \midrule
    Senso-ji Temple & temple & 500 & 1800 \\
    Shibuya Scramble Crossing & urban\_icon & 400 & 1200 \\
    Tokyo Disneyland & theme\_park & 1000 & 2400 \\
    Marunouchi Business District & business\_district & 700 & 1800 \\
    Tokyo International Forum & conference\_center & 500 & 1800 \\
    Tokyo Big Sight & conference\_center & 700 & 1800 \\
    Haneda Airport Terminal 3 & airport & 1000 & 1800 \\
    \bottomrule
    \end{tabular}
\end{table}

For a staypoint $i$ and POI $p$, the match indicator is
\begin{equation}
m_{i,p} = \mathbb{1}\!\left[d(i,p)\le r_p \ \land \ \tau_i \ge \max(\delta_p,\delta_0)\right],
\label{eq:poi_match}
\end{equation}
where $d(i,p)$ is haversine distance, $\tau_i$ is staypoint dwell time, and $\delta_0$ is a global dwell floor. If multiple POIs satisfy Eq.~\eqref{eq:poi_match}, we assign the nearest valid POI.

At the agent level, tourist identity is defined by a conjunction of episode-length, sightseeing continuity, diversity, and dwell constraints:
\begin{equation}
y^{\text{tour}}_a=\mathbb{1}\!\left[
\begin{aligned}
&2 \le e_a \le 14 \ \land\ c_a \ge 2 \ \land\ u_a \ge 3\\
&\land\ s_a \ge 2 \ \land\ h_a \ge 4 \ \land\ q_a \le 25
\end{aligned}
\right],
\label{eq:tourist_rule}
\end{equation}
where $e_a$ is episode days, $c_a$ is max consecutive sightseeing days, $u_a$ is distinct matched POIs, $s_a$ is sightseeing days, $h_a$ is total sightseeing hours, and $q_a$ is active days. To reduce ad hoc threshold choices, cutoffs are calibrated from survey marginals on nights stayed and visited wards by converting robust quantiles of the reference distributions into practical bounds. For stability, each deployed threshold is a weighted blend of the survey derived value and a hand specified default.
Surviving agents are tagged by calendar month. ward-level visit priors $P(w \mid m)$ and transition priors $P(w' \mid w, m)$ are computed from unique agent counts at the aggregate level only.

\subsection{Stage 1: Travel Parameter Prediction}

Stage 1 predicts trip extent from a compact demographic feature set. Trip scope, defined as nights stayed and number of visited destinations, is conditioned on demographic drivers including gender, standardized age, purpose of visit, total expenditure percentile, companion type, and household role. Travel purpose follows the survey's named categories (Sightseeing, Visiting relatives, Business, International conference, Expo/trade fair, Corporate conference, Incentive/Study abroad, and Other), with invalid responses removed during preprocessing. This compact feature set is designed to retain the primary behavioral drivers of trip scope while remaining applicable across destinations and survey settings.

Two independent XGBoost multi-class classifiers~\cite{chen2016xgboost} are trained for $p_{\text{nights}}$ and $p_{\text{locs}}$, each with a softmax objective. This discrete formulation directly learns $P(y|\mathbf{x})$ over feasible count values and avoids post-hoc rounding artifacts that arise in continuous regression settings.

Because pure classifier probabilities may underrepresent tail behavior, final sampling uses a blended distribution:
\begin{equation}
  P_{\text{blend}}(y|\mathbf{x}) = \alpha\, P_{\text{model}}(y|\mathbf{x}) + (1-\alpha)\, P_{\text{prior}}(y|\text{bucket}(\mathbf{x}))
  \label{eq:blend1}
\end{equation}
where $\alpha \approx 0.7$ and $\text{bucket}(\mathbf{x})$ stratifies the demographic space by purpose, gender, coarse age band, and spending quartile. This blending preserves individual conditioning while anchoring outputs to empirically plausible distributional structure.

\subsection{Stage 2: GPS-Conditioned Ward Sequence Assignment}

Stage 2 uses a two step assignment strategy that separates marginal calibration from route ordering.
First, the Stage 1 location count prediction $\hat{p}^{\text{locs}}_i$ is converted to a ward-level unique-ward budget via Eq.~\eqref{eq:unique_ward_budget}:
\begin{equation}
u_i=\text{clip}\!\left(\text{round}(\rho \,\hat{p}^{\text{locs}}_i),u_{\min},u_{\max}\right),
\label{eq:unique_ward_budget}
\end{equation}
where $\rho$ is a location-to-ward conversion ratio and $(u_{\min},u_{\max})$ are practical lower and upper bounds.
Given survey ward marginals $\pi(w)$, we allocate integer ward quotas and assign each traveler a ward set of size $u_i$. This step enforces global ward-share alignment by design.

Second, we order wards within each assigned set using GPS-derived transition structure and spatial feasibility. The month-conditioned GPS prior $P_{\text{gps}}(w'|w,m)$ is blended with a pooled fallback $P_{\text{pooled}}(w'|w)$ to regularize sparse months:
\begin{equation}
T(w'|w)=(1-\gamma)\,P_{\text{gps}}(w'|w,m)+\gamma\,P_{\text{pooled}}(w'|w),
\label{eq:trans_blend}
\end{equation}
where $\gamma\in[0,1]$ is the pooling weight controlling regularization strength. Candidate next wards are then ranked by the composite score in Eq.~\eqref{eq:score}, which combines the blended transition prior $T$ from Eq.~\eqref{eq:trans_blend}, a distance penalty, the survey marginal $\pi(w)$, and a novelty term $\text{nov}(w)$ that discourages revisiting the same ward:
\begin{align}
\text{score}(w) &= \lambda_t T(w|w_{\text{prev}})
+ \lambda_d e^{-d(w_{\text{prev}},w)/\tau} \nonumber\\
&\quad + \lambda_p \pi(w) + \lambda_n \text{nov}(w),
\label{eq:score}
\end{align}
where $d(w, w')$ is the inter-ward distance, $\tau$ is a distance decay constant, and $\lambda_t, \lambda_d, \lambda_p, \lambda_n$ are non-negative weights. The ordered route is then split across $\hat{p}^{\text{nights}}_i+1$ days to obtain the day-level ward itinerary.

\subsection{Stage 3: LLM-Based Activity Chain Generation}

In Stage 3, an LLM generates quarter-hour activity chains via API calls. Prompts encode traveler profile, Stage~1 predictions, day-level ward sequences with distance hints from the distance matrix, and a 15-type activity taxonomy adapted from NHTS categories~\cite{mcguckin2018summary}, as listed in Table~\ref{tab:activity_types}.

\begin{table}[h]
\centering
\caption{Activity type used in Stage-3 generation.}
\label{tab:activity_types}
\scriptsize
\begin{tabular}{clclcl}
\toprule
ID & Type & ID & Type & ID & Type \\
\midrule
1 & Home (Stay) & 6 & Services & 11 & Social Visits \\
2 & Work & 7 & Dining & 12 & Healthcare \\
3 & Education & 8 & Personal Errands & 13 & Religious \\
4 & Care & 9 & Recreation/Sightseeing & 14 & Miscellaneous \\
5 & Shopping & 10 & Exercise/Sports & 15 & Transport \\
\bottomrule
\end{tabular}
\end{table}

Household handling depends on travel purpose. For non-business trips, companions copy the head chain directly to represent synchronized group behavior. When a household member's purpose differs from a business-oriented head, the LLM is instructed to generate the companion chain with constrained variation, maintaining the same ward order and day count while allowing activity-type divergence to reflect the member's own purpose.

\begin{figure}[h]
    \centering
    \includegraphics[width=\columnwidth]{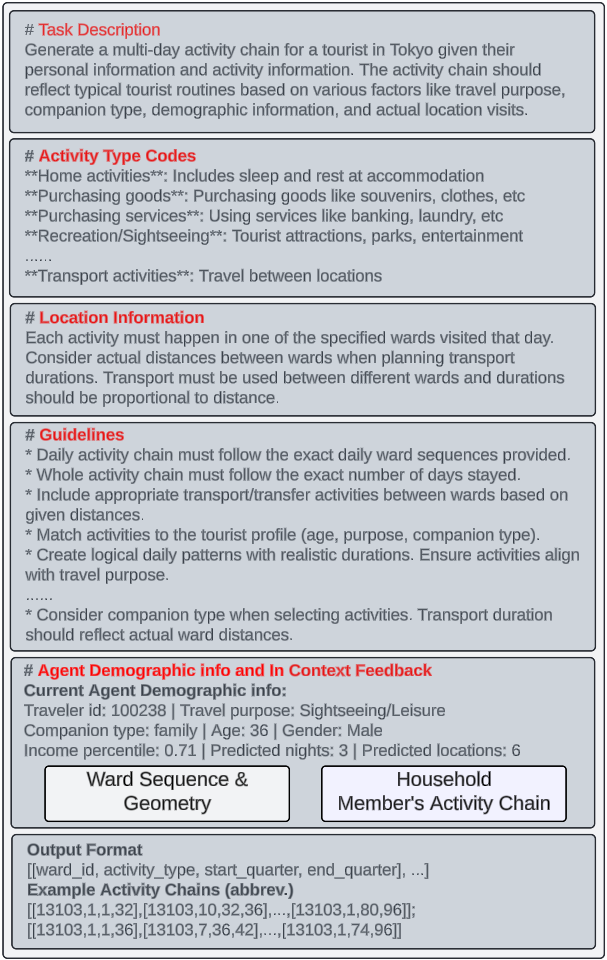}
    \caption{Structure of the Stage-3 prompt design.}
    \label{fig:prompt_structure}
\end{figure}

The generation module uses a structured prompt system illustrated in Fig.~\ref{fig:prompt_structure}. The prompt integrates the following components:
\begin{itemize}
    \item \textbf{Task description:} Defines the generation goal, instructing the LLM to produce realistic multi-day tourist activity chains conditioned on demographic and spatial inputs.
    \item \textbf{Activity type codes:} Supplies the controlled taxonomy in Table~\ref{tab:activity_types}, including numeric codes and natural-language descriptions of each activity type, to anchor generated episodes within the modeled activity space and prevent hallucinated categories.
    \item \textbf{Location information:} Provides day-level ward assignments and distance context to ensure spatial feasibility of the generated schedule.
    \item \textbf{Generation guidelines:} Specifies realism constraints including temporal coherence, purpose consistency, transport insertion, and day-to-day diversity.
    \item \textbf{Agent context and in-context feedback:} Supplies the agent's demographic profile and household context, including the head chain as a reference for companion generation, enabling coordinated household schedule production.
    \item \textbf{Output format and examples:} Constrains the output to a structured tuple format with brief examples for consistent and parseable generation.
\end{itemize}

Generated chains are validated against hard structural constraints covering day count, spatial consistency with Stage~2 ward boundaries, full temporal coverage, and absence of out-of-vocabulary activity types or ward identifiers. Non-conforming outputs are rejected and regenerated, and any residual gaps are filled with rest episodes.

\section{EXPERIMENT}

\subsection{Dataset}
\label{sec:experiment}

Experiments use two primary data sources, both from 2024. The first is the Survey on Behavioral Characteristics of Tourists, published by the Tokyo Metropolitan Government \cite{tokyo_tourism_survey2024}. This survey captures respondent demographics (gender, age, origin), trip attributes (purpose, companions, nights stayed, expenditure), and visited destinations across Tokyo's wards, and serves as the behavioral reference for population synthesis and evaluation. The second is the Veraset GPS mobility dataset \cite{veraset2024} covering the Tokyo area. Raw trajectory data are preprocessed into staypoint records by merging consecutive pings for the same agent that are both temporally and spatially adjacent. Agents with fewer than 8 distinct staypoint locations are discarded to ensure sufficient trace coverage. The resulting staypoint records are used in Stage~0 for tourist cohort extraction and monthly ward prior estimation. For spatial comparison, we aggregate visitation at ward-level and compare normalized visit shares between survey derived references and GPS identified tourists. Stage~0 tourist cohort extraction applies a global dwell floor of $\delta_0=900$ s, and the threshold blending weight is set to $\eta=0.60$. Stage~3 activity chain generation is implemented using GPT-4o-mini with a sampling temperature of 0.8.

\subsection{Experiment and Result}

We evaluate the framework stage by stage, assessing spatial cohort fidelity, trip scope prediction accuracy, monthly ward sequence alignment, and activity chain consistency. Each stage is validated against the same survey reference used to construct the synthetic population; Stage~0 and Stage~2 spatial outputs are additionally compared against GPS extracted patterns to confirm that the GPS prior is both faithfully learned and faithfully reproduced.

\begin{figure}[h]
    \centering
    \includegraphics[width=\columnwidth]{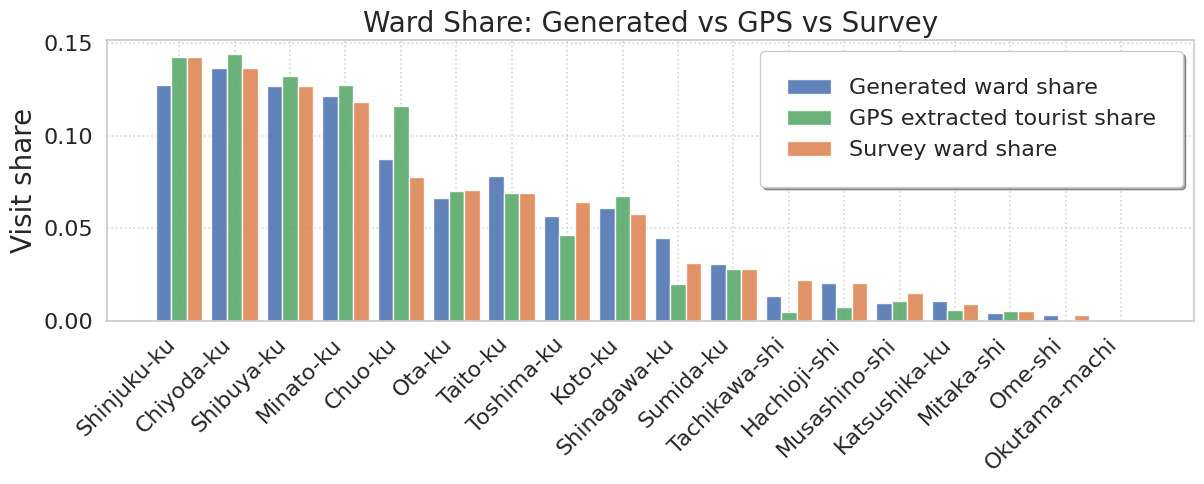}
    \caption{Ward-level visit share comparison across GPS extracted tourists, framework generated population, and survey reference. }
    \label{fig:ward_share_three}
\end{figure}

For Stage~0 and Stage~2 spatial fidelity, a critical question is whether GPS tourist cohort extraction recovers a spatial signature reflecting genuine tourist behavior rather than generic urban mobility, and whether Stage~2 propagates survey marginals into the generated population rather than drifting toward GPS overrepresented areas. Fig.~\ref{fig:ward_share_three} addresses both simultaneously. The GPS cohort, the framework generated population, and the survey reference all align closely in high traffic tourism wards, with consistent rank ordering in lower traffic wards. This three way consistency validates the Stage 0 extraction rules and confirms that the Stage 2 quota calibration successfully transfers survey spatial structure to the generated output.

\begin{figure}[h]
    \centering
    \includegraphics[width=\columnwidth]{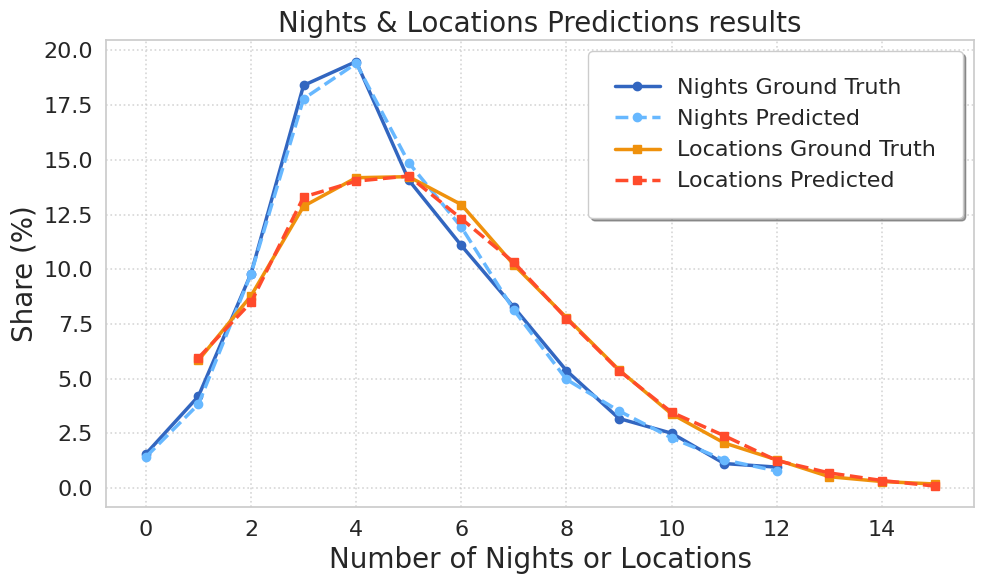}
    \caption{Stage 1 prediction results for nights stayed and number of visited locations.}
    \label{fig:nights_locs_pred}
\end{figure}

In Stage~1, trip extent, nights stayed and number of visited locations, directly governs downstream ward count and activity chain length, so it is important that the classifiers capture the full distributional spread rather than collapsing toward modal values. As shown in Fig.~\ref{fig:nights_locs_pred}, the predicted distributions closely follow survey ground truth across the full count range, including the long tails for extended stays and multi-location itineraries. The absence of systematic bias in either dimension confirms that the blended sampling strategy in Eq.~\eqref{eq:blend1} effectively retains tail behavior while respecting individual demographic conditioning.

\begin{figure}[h]
    \centering
    \includegraphics[width=\columnwidth]{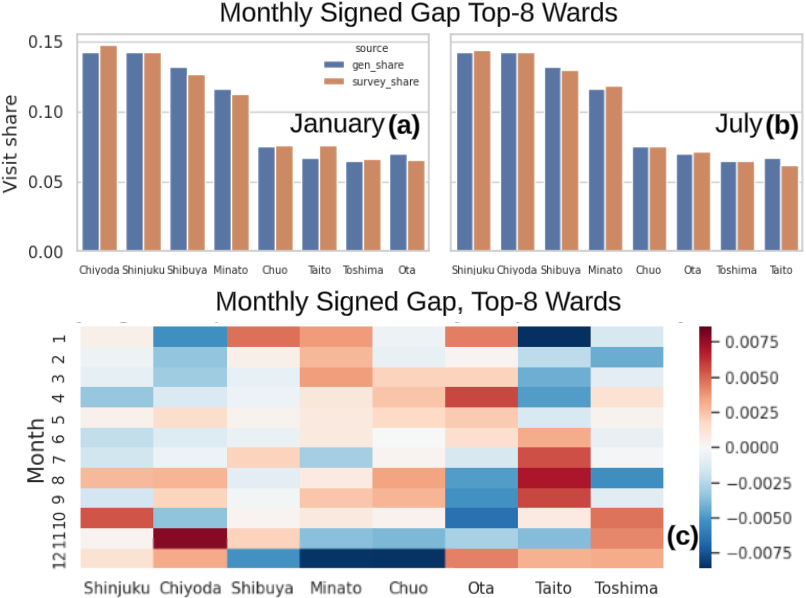}
    \caption{Monthly alignment diagnostics for Stage 2 ward shares (top-8 wards). Panel (a) shows January and panel (b) shows July generated vs survey comparisons; panel (c) shows the full month-by-ward signed gap heatmap.}
    \label{fig:monthly_gap}
\end{figure}

Month level fidelity in Stage~2 is critical because tourist demand is highly seasonal; a framework that performs well in aggregate but drifts month-by-month offers limited planning value. Fig.~\ref{fig:monthly_gap} examines January and July as representative months. Both bar charts demonstrate close agreement between generated and survey ward shares in the top-8 wards, and the signed gap heatmap ($\text{generated}-\text{survey}$) stays near zero across all twelve months with no systematic directional drift for any ward or season. This validates the month conditioned GPS transition prior design and demonstrates that the blended prior in Eq.~\eqref{eq:trans_blend} maintains seasonal fidelity without overfitting to the GPS cohort.

Beyond marginal ward shares, realistic itineraries must also capture sequential transition patterns, which wards tend to follow which. Table~\ref{tab:transition_eval} reports five complementary metrics against the GPS derived reference. A row wise JSD of 0.032 and a flow Spearman correlation of 0.884 indicate strong distributional and rank alignment. The distance Wasserstein gap of 1.881 km is modest relative to Tokyo's ward scale, and the Top-20 edge recall of 0.91 confirms that the observed GPS transitions are well covered. GPS mass coverage of 0.997 shows that virtually all GPS observed transition probability mass is represented in the generated sequences.

\begin{table}[h]
    \centering
    \caption{Stage 2 ward transition evaluation against GPS derived reference.}
    \label{tab:transition_eval}
    \scriptsize
    \begin{tabular}{@{}lccccc@{}}
    \toprule
    Metric & \makecell{Row-JSD\\$\downarrow$} & \makecell{Flow Spearman\\$\uparrow$} & \makecell{Dist. $W_1$\\(km) $\downarrow$} & \makecell{Top-20\\Recall $\uparrow$} & \makecell{GPS Mass\\Covered $\uparrow$} \\
    \midrule
    Overall   & 0.032 & 0.884 & 1.881 & 0.91 & 0.997 \\
    \bottomrule
    \end{tabular}
\end{table}

\begin{figure}[h]
    \centering
    \includegraphics[width=\columnwidth]{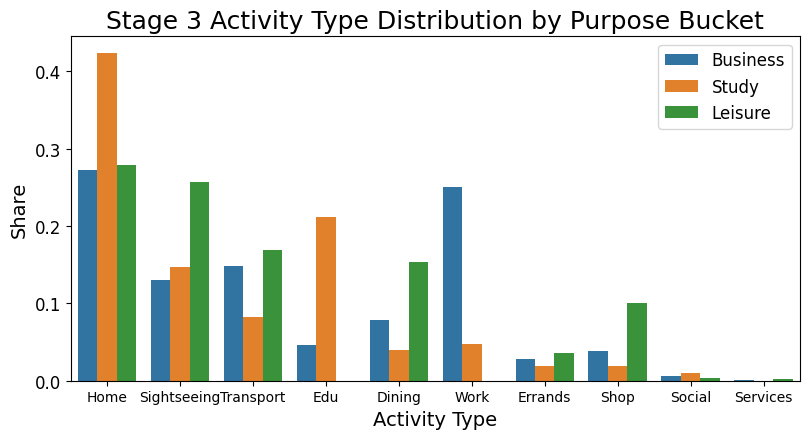}
    \caption{Stage 3 activity type distribution by purpose group.}
    \label{fig:stage3_activity_dist}
\end{figure}

Stage~3 is evaluated from two complementary angles: purpose conditioned behavioral diversity and structural consistency. For behavioral diversity, Fig.~\ref{fig:stage3_activity_dist} presents activity type distributions by purpose group. The generated chains are clearly purpose sensitive: business travelers allocate a substantially larger share to work episodes, leisure visitors show notably higher shares in sightseeing and dining, and students exhibit high education activity shares alongside elevated accommodation time. These distinctions confirm that the LLM correctly interprets the demographic and household context supplied in the prompt, rather than generating a generic activity distribution.

\begin{table}[h]
    \centering
    \caption{Stage-3 consistency diagnostics.}
    \label{tab:stage3_consistency}
    \scriptsize
    \begin{tabular}{@{}lclc@{}}
    \toprule
    Metric & Value & Metric & Value \\
    \midrule
    Day coverage rate & 98.6\% & Ward adherence rate & 92.0\% \\
    Night-count alignment rate & 95.4\% & Hallucination rate & 2.2\% \\
    \bottomrule
    \end{tabular}
\end{table}

For structural consistency, Table~\ref{tab:stage3_consistency} reports four hard constraint metrics. Day coverage of 98.6\% and night count alignment of 95.4\% confirm that the LLM correctly reproduces the trip duration specified by Stage~1. Ward adherence of 92.0\% indicates that generated episodes largely respect the Stage~2 spatial assignments, and a hallucination rate of only 2.2\% shows that out of vocabulary activity types and ward identifiers are rare, reflecting the effectiveness of the structured prompt taxonomy and post generation validation.

Taken together, the experiments indicate that the pipeline yields high quality synthetic tourist trajectories. GPS derived cohort extraction recovers ward-level visitation consistent with the survey, ward sequences align with survey marginals and GPS transition structure at both annual and monthly levels, and activity chain generation is purpose sensitive with high structural consistency. These results support the framework's viability for planning oriented tourist demand synthesis.

\section{Conclusion and Future Work}

This paper presented a four stage framework for generating demographically conditioned, spatially grounded tourist itineraries without individual level GPS traces. Experiments in Tokyo show close agreement with survey distributions and GPS derived monthly visitation and transition patterns. Future work will improve companion variation, expand POI coverage and Stage~0 sensitivity analysis, test transferability to other destinations, and incorporate contextual signals such as weather, events, and day-of-week effects.


\section*{Acknowledgment}
This work was supported by the Intelligence Advanced Research Projects Activity (IARPA) via the Department of Interior/Interior Business Center (DOI/IBC) contract number 140D0423C0033. The U.S. Government is authorized to reproduce and distribute reprints for Governmental purposes notwithstanding any copyright annotation thereon. Disclaimer: The views and conclusions contained herein are those of the authors and should not be interpreted as necessarily representing the official policies or endorsements, either expressed or implied, of IARPA, DOI/IBC, or the U.S. Government.

\bibliographystyle{IEEEtran}
\bibliography{itsc_2026_tourist_model}

\end{document}